\documentclass[11pt,a4paper]{article}

\usepackage[hyperref]{emnlp2018}
\usepackage{times}
\usepackage{latexsym}
\usepackage{amsfonts}
\usepackage{amsmath}
\usepackage{algorithm}

\usepackage{multirow} 
\usepackage{colortbl} 

\usepackage{url}

\usepackage[utf8]{inputenc} 
\usepackage{textcomp}

\usepackage[hyperref]{emnlp2018}
\usepackage{times}
\usepackage{latexsym}
\usepackage{todos}

\usepackage{amsmath}
\usepackage{amssymb}

\usepackage{bm}
\usepackage{color}
\usepackage{multirow}
\usepackage{graphicx, wrapfig}
\usepackage[font=normalsize,position=bottom]{subfig}
\usepackage{arydshln}
\usepackage{balance}

\aclfinalcopy

\begin{document}

\title{Exploiting Deep Representations for Neural Machine Translation}

\author{Zi-Yi Dou\\\normalsize Nanjing University\\{\normalsize \tt douzy@smail.nju.edu.cn} \And
Zhaopeng Tu\thanks{~~~Zhaopeng Tu is the corresponding author of the paper. This work was conducted when Zi-Yi Dou was interning at Tencent AI Lab.}\\\normalsize Tencent AI Lab\\{\normalsize \tt zptu@tencent.com} \And
Xing Wang\\\normalsize Tencent AI Lab\\{\normalsize \tt brightxwang@tencent.com} \AND
Shuming Shi\\\normalsize Tencent AI Lab\\{\normalsize \tt shumingshi@tencent.com} \And
Tong Zhang\\\normalsize Tencent AI Lab\\{\normalsize \tt bradymzhang@tencent.com}
}

\maketitle

\begin{abstract}
%Neural machine translation (NMT) systems have demonstrated their effectiveness and capacity in many language pairs. Deep neural networks have attracted huge attention from researchers and have been widely applied in NMT because of their ability to model complex functions and capture complicated linguistic structures. Typically, state-of-the-art NMT systems have encoder and decoder with multiple layers. Although deep architecture have proven to be effective and useful, we argue that one common drawback of these models is that they only utilize the information in the top layer. Recently, several studies have indicated that different layers can capture diverse information. In this paper, we propose several strategies to combine the information across layers, from simple linear combination to deep, nonlinear layer aggregation. By augmenting the state-of-the-art architecture with deeper aggregation, we manage to learn better representations and get better translation quality for NMT systems. Experimental results on WMT14 English$\Rightarrow$German and WMT17 Chinese$\Rightarrow$English datasets demonstrate the necessity of fusing layers and the effectiveness of our proposed approaches.

Advanced neural machine translation (NMT) models generally implement encoder and decoder as multiple layers, which allows systems to model complex functions and capture complicated linguistic structures.
However, only the top layers of encoder and decoder are leveraged in the subsequent process, which misses the opportunity to exploit the useful information embedded in other layers.
In this work, we propose to simultaneously expose all of these signals with layer aggregation and multi-layer attention mechanisms. In addition, we introduce an auxiliary regularization term to encourage different layers to capture diverse information.
Experimental results on widely-used WMT14 English$\Rightarrow$German and WMT17 Chinese$\Rightarrow$English translation data demonstrate the effectiveness and universality of the proposed approach.

% However, we argue that one common drawback of these deep models is that they only utilize the information in the top layer. Inspired by the recent studies which have indicated that different layers can capture diverse information, in this paper, we propose several strategies to combine the information across layers, from simple linear combination to deep nonlinear layer aggregation. By augmenting the state-of-the-art architecture with deeper aggregation, we manage to learn better representations and get better translation quality for NMT systems. Experimental results on WMT14 English$\Rightarrow$German and WMT17 Chinese$\Rightarrow$English datasets demonstrate the necessity of fusing layers and the effectiveness of our proposed approaches.
\end{abstract}

\section{Introduction}

Neural machine translation (NMT) models have advanced the machine translation community in recent years~\cite{kalchbrenner2013recurrent,cho2014learning,sutskever2014sequence}. NMT models generally consist of two components: an {\em encoder} network to summarize the input sentence into sequential representations, based on which a {\em decoder} network generates target sentence word by word with an attention model~\cite{Bahdanau:2015:ICLR,luong2015effective}.

Nowadays, advanced NMT models generally implement encoder and decoder as multiple layers, regardless of the specific model architectures such as RNN~\cite{Zhou:2016:TACL,wu2016google}, CNN~\cite{Gehring:2017:ICML}, or Self-Attention Network~\cite{Vaswani:2017:NIPS,chen2018the}.
Several researchers have revealed that different layers are able to capture different types of syntax and semantic information~\cite{shi-padhi-knight:2016:EMNLP2016,Peters:2018:NAACL,Anastasopoulos:2018:NAACL}. For example,~\newcite{shi-padhi-knight:2016:EMNLP2016} find that both local and global source syntax are learned by the NMT encoder and different types of syntax are captured at different layers.

However, current NMT models only leverage the top layers of encoder and decoder in the subsequent process, which misses the opportunity to exploit useful information embedded in other layers.
Recently, aggregating layers to better fuse semantic and spatial information has proven to be of profound value in computer vision tasks~\cite{Huang:2017:CVPR,Yu:2018:CVPR}.
In natural language processing community,~\newcite{Peters:2018:NAACL} have proven that simultaneously exposing all layer representations outperforms methods that utilize just the top layer for transfer learning tasks. 

Inspired by these findings, we propose to exploit deep representations for NMT models. Specifically, we investigate two types of strategies to better fuse information across layers, ranging from layer aggregation to multi-layer attention. While layer aggregation strategies combine hidden states at the same position across different layers, multi-layer attention allows the model to combine information in different positions. In addition, we introduce an auxiliary objective to encourage different layers to capture diverse information, which we believe would make the deep representations more meaningful.

We evaluated our approach on two widely-used WMT14 English$\Rightarrow$German and WMT17 Chinese$\Rightarrow$English translation tasks. We employed \textsc{Transformer}~\cite{Vaswani:2017:NIPS} as the baseline system since it has proven to outperform other architectures on the two tasks~\cite{Vaswani:2017:NIPS,hassan2018achieving}.
Experimental results show that exploiting deep representations consistently improves translation performance over the vanilla \textsc{Transformer} model across language pairs. 
It is worth mentioning that \textsc{Transformer-Base} with deep representations exploitation outperforms the vanilla \textsc{Transformer-Big} model with only less than half of the parameters.% and less training time.

\section{Background: Deep NMT}

Deep representations have proven to be of profound value in machine translation~\cite{Meng:2016:ICLRWorkshop,Zhou:2016:TACL}. Multiple-layer encoder and decoder are employed to perform the translation task through a series of nonlinear transformations from the representation of input sequences to final output sequences.
The layer can be implemented as RNN~\cite{wu2016google}, CNN~\cite{Gehring:2017:ICML}, or Self-Attention Network~\cite{Vaswani:2017:NIPS}. In this work, we take the advanced Transformer as an example, which will be used in experiments later.
However, we note that the proposed approach is generally applicable to any other type of NMT architectures.

Specifically, the encoder is composed of a stack of $L$ identical layers, each of which has two sub-layers. The first sub-layer is a self-attention network, and the second one is a position-wise fully connected feed-forward network. A residual connection~\cite{he2016deep} is employed around each of the two sub-layers, followed by layer normalization~\cite{ba2016layer}.
Formally, the output of the first sub-layer ${\bf C}_e^l$ and the second sub-layer ${\bf H}_e^l$ are calculated as
\begin{eqnarray}
    {\bf C}_e^l &=& \textsc{Ln}\big(\textsc{Att}({\bf Q}_e^{l}, {\bf K}_e^{l-1}, {\bf V}_e^{l-1}) + {\bf H}_e^{l-1} \big), \nonumber\\
    {\bf H}_e^l &=& \textsc{Ln}\big(\textsc{Ffn}({\bf C}_e^l) + {\bf C}_e^{l} \big),
    \label{eqn:enc}
\end{eqnarray}
where $\textsc{Att}(\cdot)$, $\textsc{Ln}(\cdot)$, and $\textsc{Ffn}(\cdot)$ are self-attention mechanism, layer normalization, and feed-forward networks with ReLU activation in between, respectively. $\{{\bf Q}_e^l, {\bf K}_e^{l-1}, {\bf V}_e^{l-1}\}$ are query, key and value vectors that are transformed from the ({\em l-1})-th encoder layer ${\bf H}_e^{l-1}$.

The decoder is also composed of a stack of $L$ identical layers. In addition to two sub-layers in each decoder layer, the decoder inserts a third sub-layer ${\bf D}_d^l$ to perform attention over the output of the encoder stack ${\bf H}^{L}_{e}$:
\begin{eqnarray}
    {\bf C}_d^l &=& \textsc{Ln}\big(\textsc{Att}({\bf Q}_d^{l}, {\bf K}_d^{l-1}, {\bf V}_d^{l-1}) + {\bf H}_d^{l-1} \big), \nonumber\\
    {\bf D}_d^l &=& \textsc{Ln}\big(\textsc{Att}({\bf C}_d^{l}, {\bf K}_e^L, {\bf V}_e^L) + {\bf C}_d^{l} \big), \nonumber \\
    {\bf H}_d^l &=& \textsc{Ln}\big(\textsc{Ffn}({\bf D}_d^l) + {\bf D}_d^{l} \big),
    \label{eqn:dec}
\end{eqnarray}
where $\{{\bf Q}_d^l, {\bf K}_d^{l-1}, {\bf V}_d^{l-1}\}$ are transformed from the ({\em l-1})-th decoder layer ${\bf H}_d^{l-1}$, and $\{{\bf K}_e^{L}, {\bf V}_e^{L}\}$ are transformed from the top layer of the encoder. The top layer of the decoder ${\bf H}_d^L$ is used to generate the final output sequence.

\vspace{5pt}

Multi-layer network can be considered as a strong feature extractor with extended receptive fields capable of linking salient features from the entire sequence~\cite{chen2018the}. However, one potential problem about the vanilla Transformer, as shown in Figure~\ref{fig:vanilla}, is that both the encoder and decoder stack layers in sequence and only utilize the information in the top layer. 
While studies have shown deeper layers extract more semantic and more global features~\cite{zeiler2014visualizing,Peters:2018:NAACL}, these do not prove that the last layer is the ultimate representation for any task. 
Although residual connections have been incorporated to combine layers, these connections have been ``shallow'' themselves, and only fuse by simple, one-step operations~\cite{Yu:2018:CVPR}. 
We investigate here how to better fuse information across layers for NMT models.

In the following sections, we simplify the equations to ${\bf H}^l = \textsc{Layer}({\bf H}^{l-1})$ for brevity.

\section{Proposed Approaches}

In this section, we first introduce how to exploit deep representations by simultaneously exposing all of the signals from all layers (Sec~\ref{sec-deep-representations}).
% two types of mechanisms to fuse information across layers, from layer aggregation to multi-layer attention (Sec~\ref{sec-deep-representations}). 
% the proposed layer aggregation strategies to fuse information across layers, from simple linear combination to deep, non-linear aggregation. 
Then, to explicitly encourage different layers to incorporate various information, we propose one way to measure the diversity between layers and add a regularization term to our objective function to maximize the diversity across layers (Sec~\ref{sec:div}). 
% Finally, we illustrate our training procedure and objective.

\subsection{Deep Representations}
\label{sec-deep-representations}

\begin{figure}[t]
\centering
\subfloat[Vanilla]{
\includegraphics[width=0.12\textwidth]{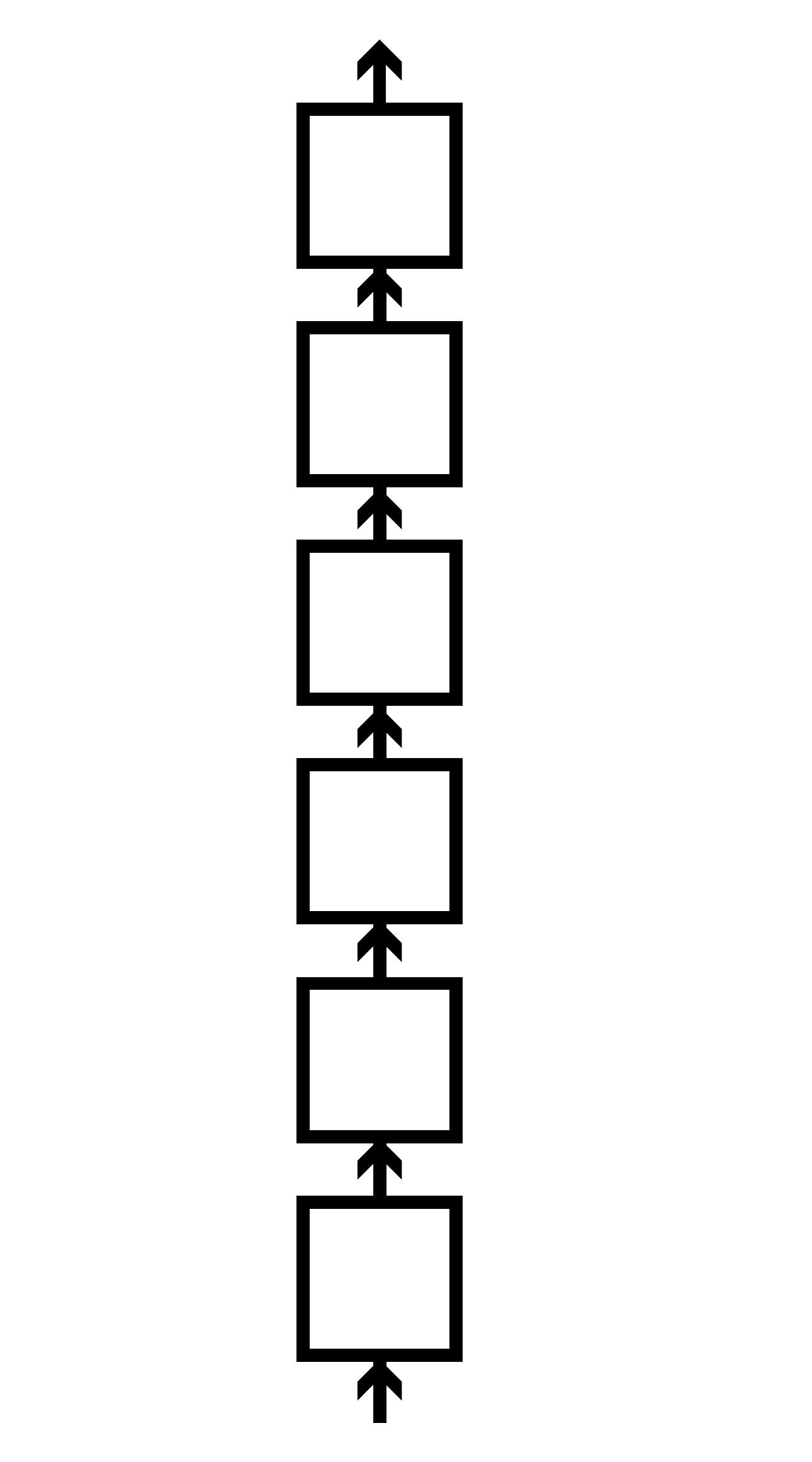}
\label{fig:vanilla}
} \hspace{0.1\textwidth}
\subfloat[Dense]{
\includegraphics[width=0.12\textwidth]{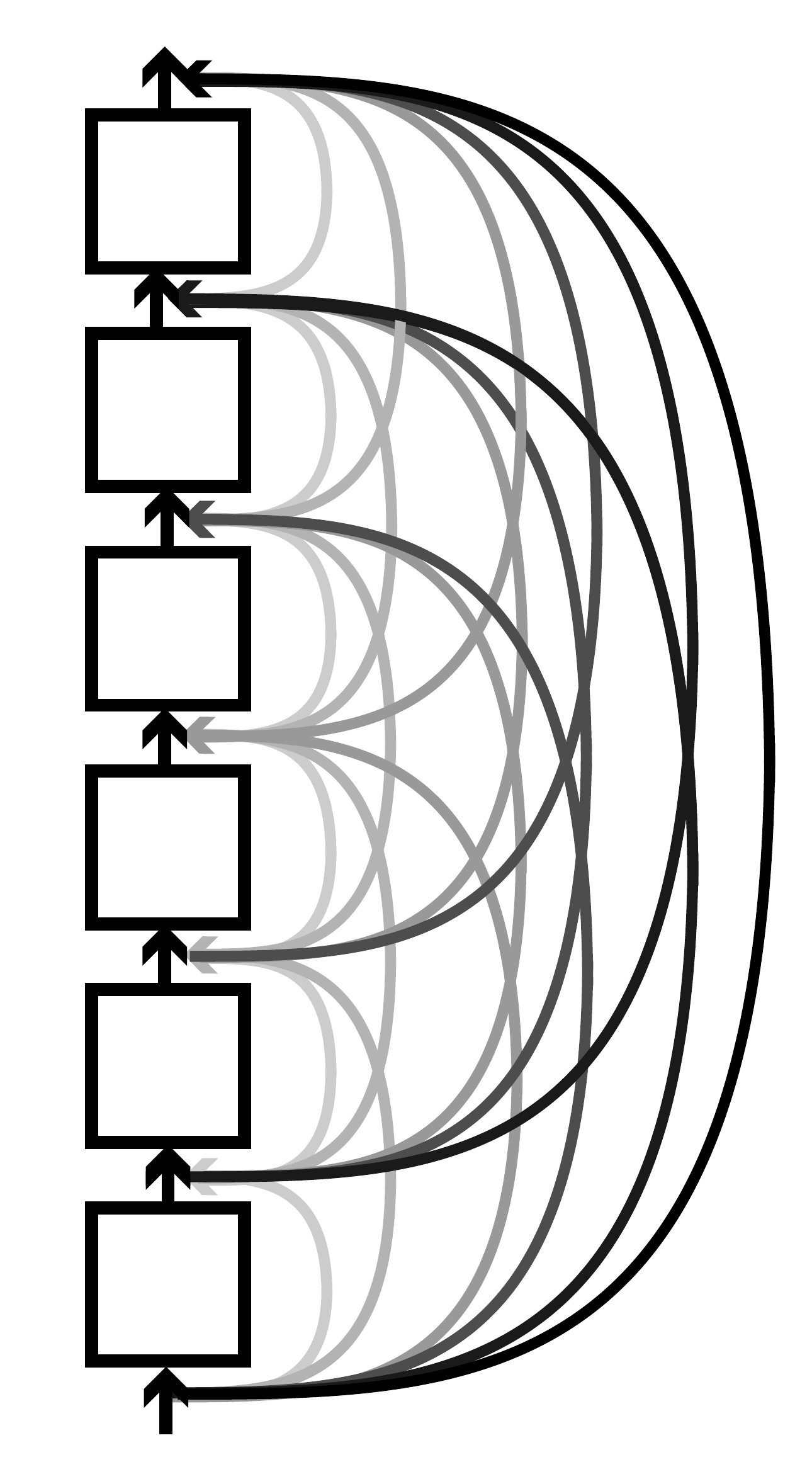}
\label{fig:dense}
} \\
\subfloat[Linear]{
\includegraphics[width=0.12\textwidth]{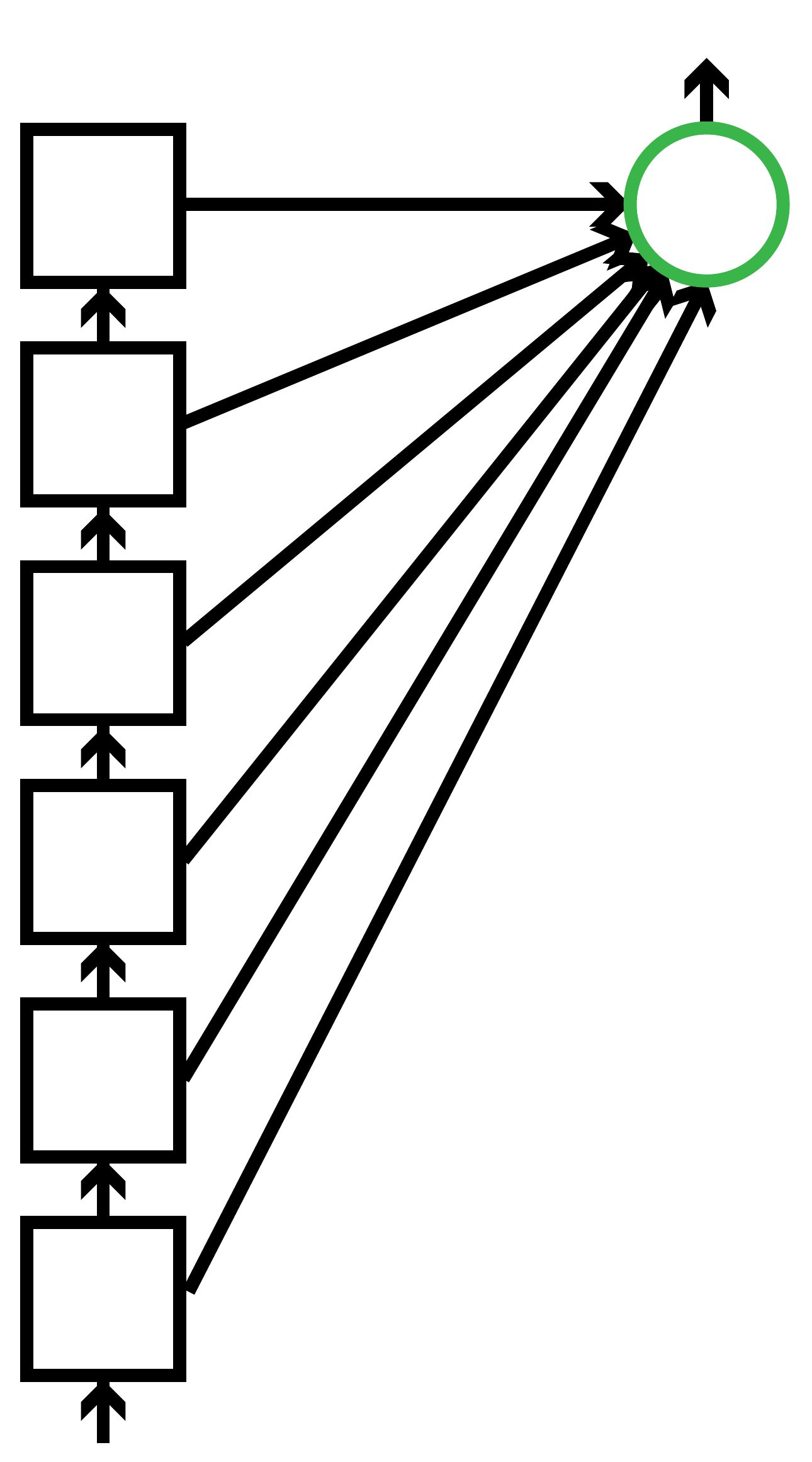}
\label{fig:linear}
} \hspace{0.1\textwidth}
\subfloat[Iterative]{
\includegraphics[width=0.12\textwidth]{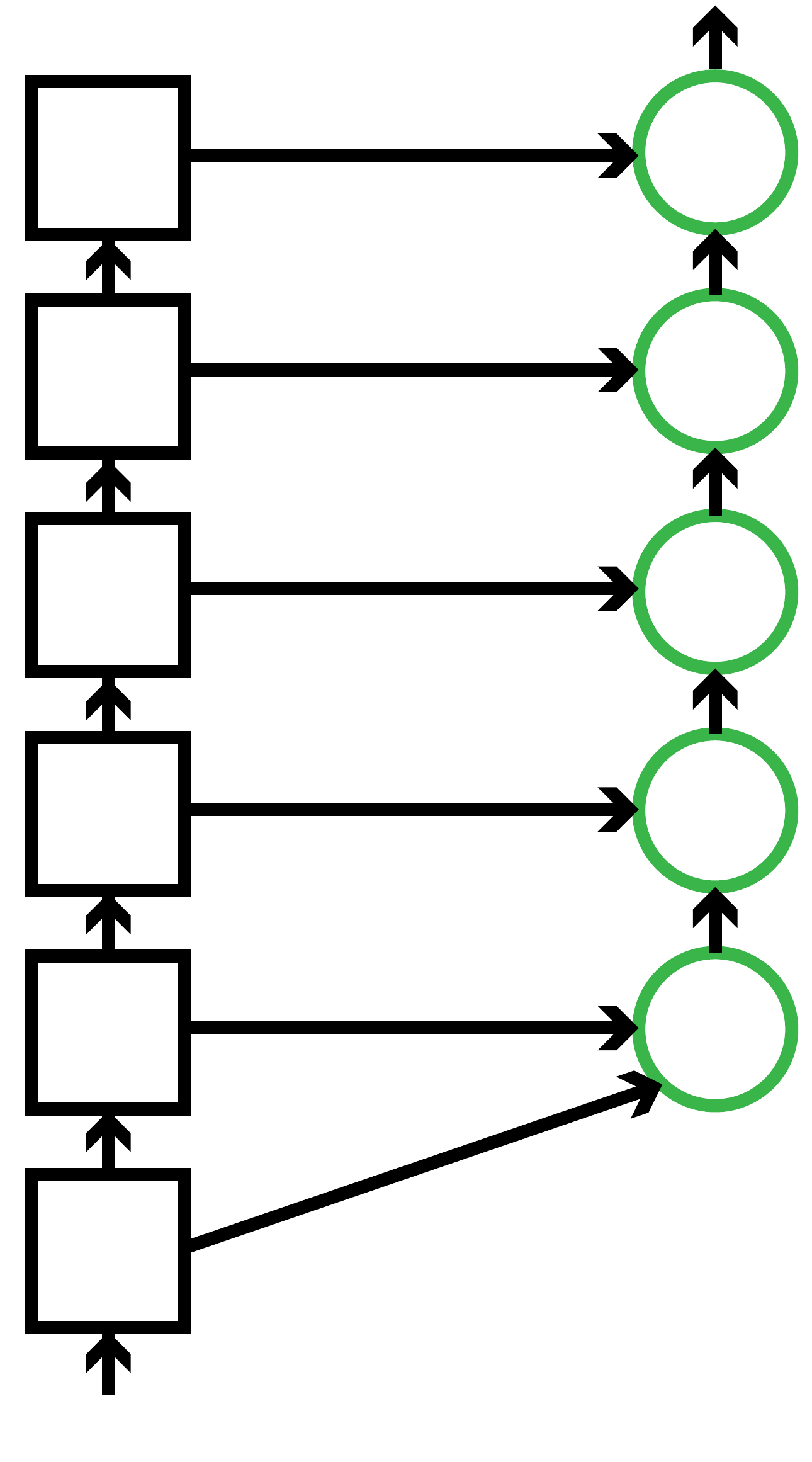}
\label{fig:ite}
}
\caption{Illustration of (a) vanilla model without any aggregation and (b,c,d) layer aggregation strategies. Aggregation nodes are represented by green circles.}
\label{fig:model_overveiw}
\end{figure}

To exploit deep representations, we investigate two types of strategies to fuse information across layers, from layer aggregation to multi-layer attention. 
% Comparing the strategies of layer aggregation and multi-layer attention, the latter may benefit from more modeling flexibility.
While layer aggregation strategies combine hidden states at the same position across different layers, multi-layer attention allows the model to combine information in different positions.

\subsubsection{Layer Aggregation}
\label{sec:aggregation}

While the aggregation strategies are inspired by previous work, there are several differences since we have simplified and generalized from the original model, as described below.

\vspace{-5pt}
\paragraph{\bf Dense Connection.}
The first strategy is to allow all layers to directly access previous layers:
\vspace{-10pt}
\begin{eqnarray}
    {\bf H}^{l} = f({\bf H}^1, \dots, {\bf H}^{l-1}).
\end{eqnarray}
In this work, we mainly investigate whether densely connected networks work for NMT, which have proven successful in computer vision tasks~\cite{Huang:2017:CVPR}.
% To this end, we exploit several aggregation strategies proposed by previous work:
%Can we place direct connections between lower-level layers and the highest-level layer? We expect the highest-level layer, which is used for querying by the decoder, can more easily embed information from lower-level layers. 
The basic strategy of densely connected networks is to connect each layer to every previous layer with a residual connection:
  \begin{equation}
      \mathbf{H}^l = \text{Layer}({\bf H}^{l-1}) + \sum_{i=1}^{l-1} \mathbf{H}^i.
  \end{equation}
Figure \ref{fig:dense} illustrates the idea of this approach. Our implementation differs from~\cite{Huang:2017:CVPR} in that we use an addition instead of a concatenation operation in order to keep the state size constant. Another reason is that concatenation operation is computationally expensive, while residual connections are more efficient.

\vspace{10pt}
While dense connection directly feeds previous layers to the subsequent layers, the following mechanisms maintain additional layers to aggregate standard layers, from shallow linear combination, to deep non-linear aggregation.

\paragraph{\bf Linear Combination.}
As shown in Figure \ref{fig:linear}, an intuitive strategy is to linearly combine the outputs of all layers:
\begin{equation}
    \widehat{\mathbf{H}} = \sum_{l=1}^L {\bf W}_l \mathbf{H}^l,      
\end{equation}
where $\{{\bf W}_1, \dots, {\bf W}_L\}$ are trainable matrices.
While the strategy is similar in spirit to~\cite{Peters:2018:NAACL}, there are two main differences: (1) they use normalized weights while we directly use parameters that could be either positive or negative numbers, which may benefit from more modeling flexibility.
(2) they use a scalar that is shared by all elements in the layer states, while we use learnable matrices. The latter offers a more precise control of the combination by allowing the model to be more expressive than scalars~\cite{Tu:2017:TACL}.

\vspace{10pt}

We also investigate strategies that {\em iteratively} and {\em hierarchically} merge layers by incorporating more depth and sharing, which have proven effective for computer vision tasks~\cite{Yu:2018:CVPR}.
  
\paragraph{\bf Iterative Aggregation.} 
As illustrated in Figure \ref{fig:ite}, iterative aggregation follows the iterated stacking of the backbone architecture. Aggregation begins at the shallowest, smallest scale and then iteratively merges deeper, larger scales. The iterative deep aggregation function $I$ for a series of layers $\mathbf{H}_1^l = \{\mathbf{H}^1, \cdots, \mathbf{H}^l\}$ with increasingly deeper and semantic information is formulated as
  \begin{eqnarray}
        \widehat{\bf H}^l = I(\mathbf{H}_1^l) = \textsc{Agg}(\mathbf{H}^l, \widehat{\mathbf{H}}^{l-1}),
%                        &=& \textsc{Ln}(\textsc{Ffn}([\mathbf{H}^l, I(\mathbf{H}_1^{l-1})]) + \mathbf{H}^l +  I(\mathbf{H}_1^{l-1})
  \end{eqnarray}
where we set $ \widehat{\bf H}^1 = {\bf H}^1$ and $\textsc{Agg}(\cdot, \cdot)$ is the aggregation function:
\begin{equation}
      \textsc{Agg}(x,y) = \textsc{Ln}(\textsc{Ffn}([x;y]) + x + y).
\label{eqn-agg}
\end{equation}
As seen, in this work, we first concatenate $x$ and $y$ into $z = [x; y]$, which is subsequently fed to a feed-forward network with a sigmoid activation in between.
%$\text{FF}$ consists of two linear transformation with a sigmoid activation in between. 
Residual connection and layer normalization are also employed. Specifically, both $x$ and $y$ have residual connections to the output.
The choice of the aggregation function will be further studied in the experiment section.

\begin{figure}[t]
\centering
\subfloat[CNN-like tree]{
\includegraphics[width=0.18\textwidth]{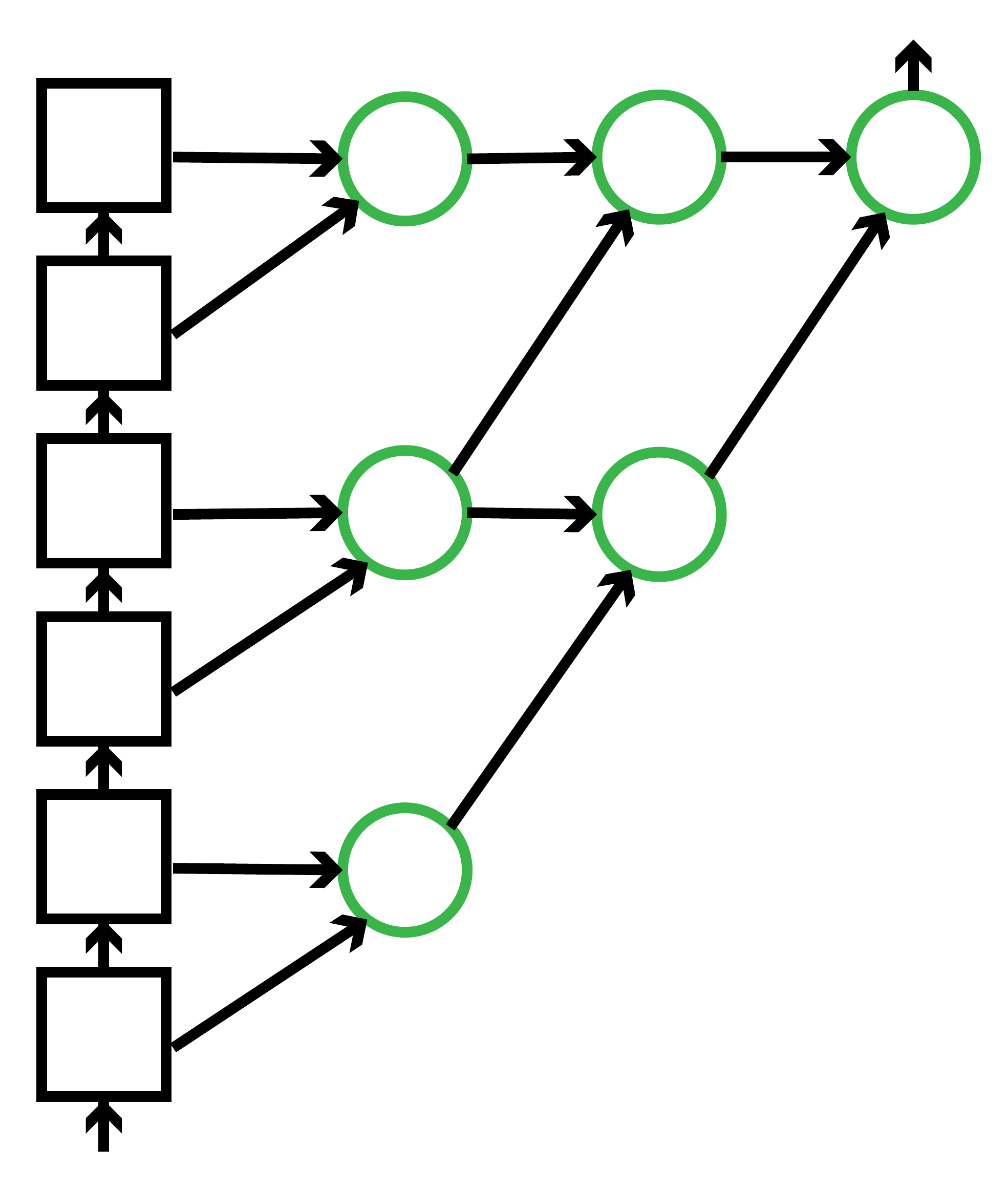}
\label{fig:tree}
} \hfill
\subfloat[Hierarchical]{
\includegraphics[width=0.18\textwidth]{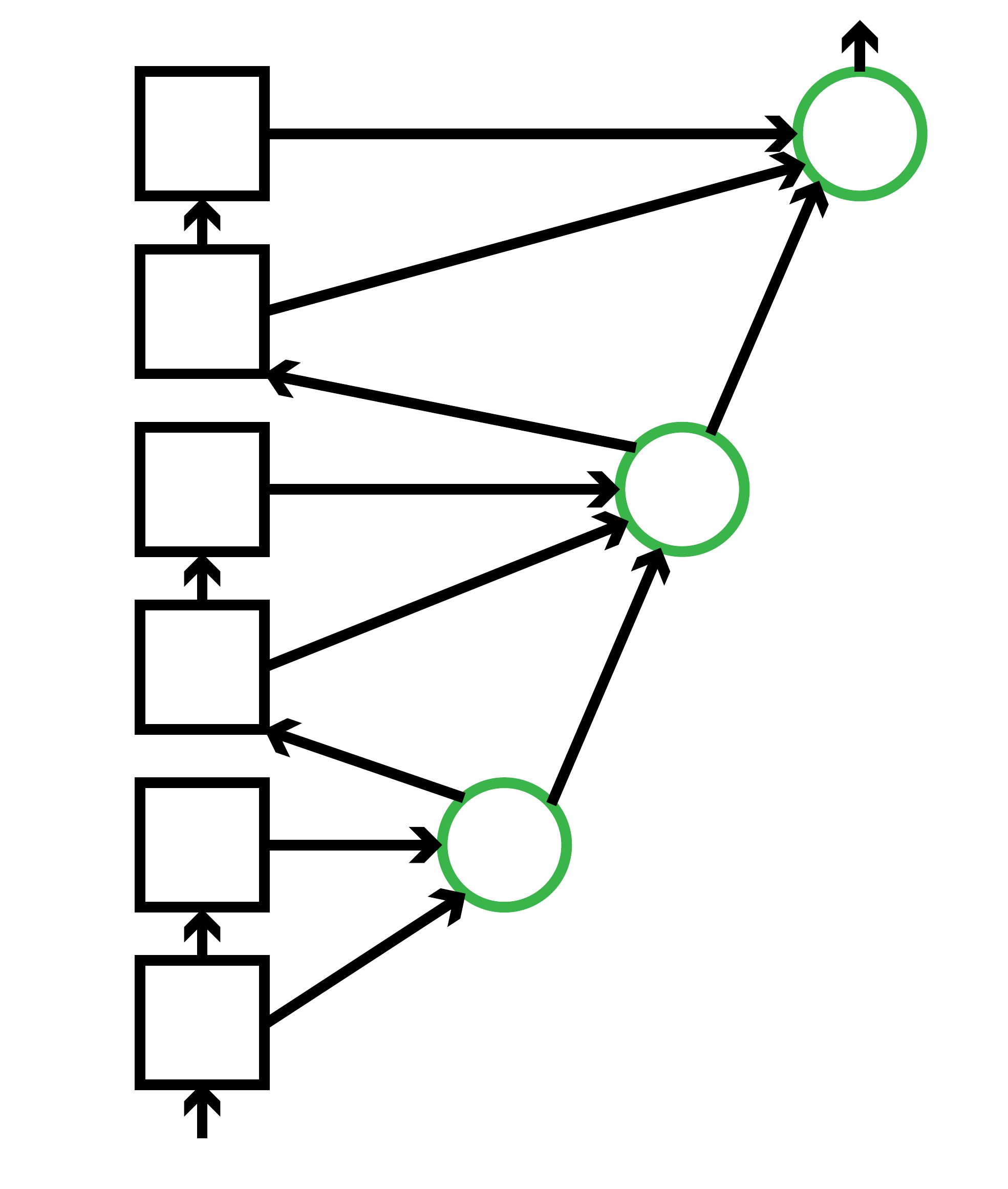}
\label{fig:hie3}
}
\caption{Hierarchical aggregation (b) that aggregates layers through a tree structure (a).}
\label{fig:tree-hier}
\end{figure}  
  
\paragraph{\bf Hierarchical Aggregation.}

While iterative aggregation deeply combines states, it may still be insufficient to fuse the layers for its sequential architecture. Hierarchical aggregation, on the other hand, merges layers through a tree structure to preserve and combine feature channels, as shown in Figure \ref{fig:tree-hier}. The original model proposed by~\newcite{Yu:2018:CVPR} requires the number of layers to be the power of two, which limits the applicability of these methods to a broader range of NMT architectures (e.g. six layers in~\cite{Vaswani:2017:NIPS}). To solve this problem, we introduce a CNN-like tree with the filter size being two, as shown in Figure~\ref{fig:tree}. Following~\cite{Yu:2018:CVPR}, we first merge aggregation nodes of the same depth for efficiency so that there would be at most one aggregation node for each depth. Then, we further feed the output of an aggregation node back into the backbone as the input to the next sub-tree, instead of only routing intermediate aggregations further up the tree, as shown in Figure~\ref{fig:hie3}. The interaction between aggregation and backbone nodes allows the model to better preserve features.

Formally, each aggregation node $\widehat{\bf H}^i$ is calculated as
\begin{align}
\widehat{\bf H}^i=&
\begin {cases}
\textsc{Agg}({\bf H}^{2i-1}, {\bf H}^{2i}), \quad\quad\quad\quad  i = 1 \\
\textsc{Agg}({\bf H}^{2i-1}, {\bf H}^{2i}, \widehat{\bf H}^{i-1}), ~~~~~ i>1 \nonumber %=2, 3 \nonumber
\end{cases}
\end{align}
where $\textsc{Agg}({\bf H}^{2i-1}, {\bf H}^{2i})$ is computed via Eqn.~\ref{eqn-agg}, and $\textsc{Agg}({\bf H}^{2i-1}, {\bf H}^{2i}, \widehat{\bf H}^{i-1})$ is computed as
\begin{equation}
      \textsc{Agg}(x,y, z) = \textsc{Ln}(\textsc{Ffn}([x;y;z]) + x + y + z). \nonumber
\label{eqn-agg-more}
\end{equation}
The aggregation node at the top layer $\widehat{\bf H}^{L/2}$ serves as the final output of the network.

\subsubsection{Multi-Layer Attention}

\begin{figure}[t]
\centering
\includegraphics[width=0.35\textwidth]{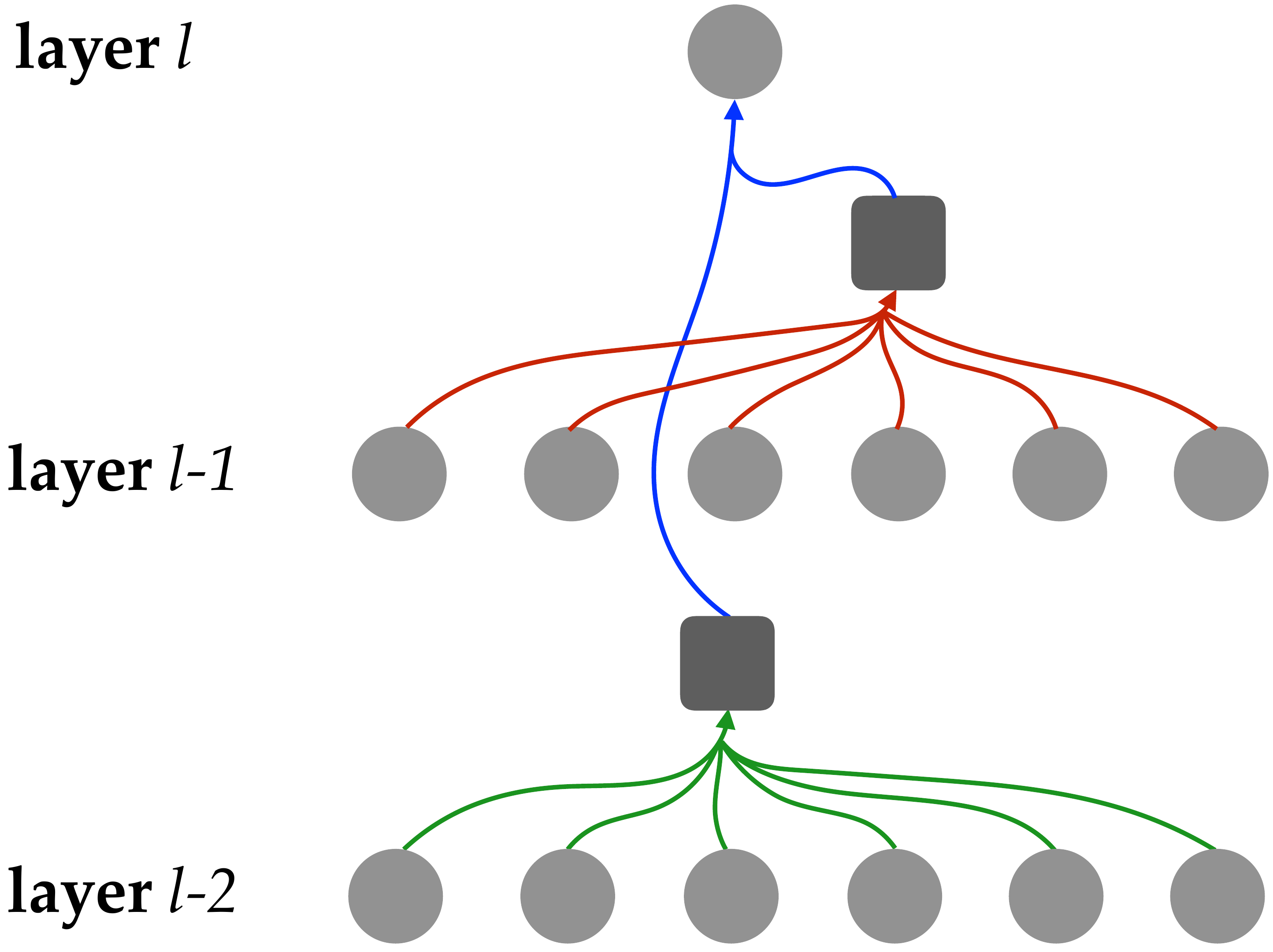}
\caption{Multi-layer attention allows the model to attend multiple layers to construct each hidden state. We use two-layer attention for illustration, while the approach is applicable to any layers lower than $l$.}
\label{fig:attention}
\end{figure}

Partially inspired by~\newcite{Meng:2016:ICLRWorkshop}, we also propose to introduce a multi-layer attention mechanism into deep NMT models, for more power of transforming information across layers. In other words, for constructing each hidden state in any layer-$l$, we allow the self-attention model to attend any layers lower than $l$, instead of just layer {\em l-1}:
\begin{eqnarray}
{\bf C}^{l}_{-1} &=& \textsc{Att}({\bf Q}^{l}, {\bf K}^{l-1}, {\bf V}^{l-1}), \nonumber\\
{\bf C}^{l}_{-2} &=& \textsc{Att}({\bf Q}^{l}, {\bf K}^{l-2}, {\bf V}^{l-2}), \nonumber\\
\dots \nonumber \\
{\bf C}^{l}_{-k} &=& \textsc{Att}({\bf Q}^{l}, {\bf K}^{l-k}, {\bf V}^{l-k}), \nonumber\\
{\bf C}^l &=& \textsc{Agg}({\bf C}^{l}_{-1}, \dots, {\bf C}^{l}_{-k}),
\end{eqnarray}
where ${\bf C}^{l}_{-i}$ is sequential vectors queried from layer {\em l-i} using a separate attention model, and
$\textsc{Agg}(\cdot)$ is similar to the pre-defined aggregation function to transform $k$ vectors $\{{\bf C}^{l}_{-1}, \dots, {\bf C}^{l}_{-k}\}$ to a $d$-dimension vector, which is subsequently used to construct the encoder and decoder layers via Eqn.~\ref{eqn:enc} and~\ref{eqn:dec} respectively. Note that multi-layer attention only modifies the self-attention blocks in both encoder and decoder, while does not revises the encoder-decoder attention blocks.

\subsection{Layer Diversity}
\label{sec:div}

Intuitively, combining layers would be more meaningful if different layers are able to capture diverse information. Therefore, we explicitly add a {\em regularization} term to encourage the diversities between layers:
\begin{equation}
    \mathcal{L} = \mathcal{L}_{\text{likelihood}} + \lambda \mathcal{L}_{\text{diversity}},
\end{equation}
where $\lambda$ is a hyper-parameter and is set to $1.0$ in this paper.
Specifically, the regularization term measures the average of the distance between every two adjacent layers:
\begin{equation}
    \mathcal{L}_{diversity} = \frac{1}{L-1}\sum_{l=1}^{L-1} D({\bf H}^{l}, {\bf H}^{l+1}).
\end{equation}
Here $D({\bf H}^{l},{\bf H}^{l+1})$ is the averaged {\em cosine-squared distance} between the states in layers 
${\bf H}^{l}=\{{\bf h}^l_1, \dots, {\bf h}^l_N\}$ and ${\bf H}^{l+1}=\{{\bf h}^{l+1}_1, \dots, {\bf h}^{l+1}_N\}$:
\begin{equation}
    D({\bf H}^{l}, {\bf H}^{l+1}) = \frac{1}{N}\sum_{n=1}^N (1 - \cos^2 ({\bf h}^l_n, {\bf h}^{l+1}_n)). \nonumber
\end{equation}
The cosine-squared distance between two vectors is maximized when two vectors are linearly independent and minimized when two vectors are linearly dependent, which satisfies our goal.\footnote{We use cosine-squared distance instead of cosine distance, since the latter is maximized when two vectors are in opposite directions. In such case, the two vectors are in fact linearly dependent, while we aim at encouraging the vectors independent from each other.}

\section{Experiments}

\begin{table*}[t]
  \centering
  \renewcommand{\arraystretch}{1.1}
  \begin{tabular}{c|l||r r r||c|c}
    {\bf \#}    &   {\bf Model} &  \bf {\# Para.} & \bf {Train}  &   \bf Decode  &    \bf  BLEU  &   $\bigtriangleup$ \\
    \hline
    1   &   \textsc{Transformer-Base} 			  & $88.0$M	&   $1.82$  & $1.33$ &  $27.64$   &   --\\
    \hline
    2   &   ~~ + Dense Connection				          & $+0.0$M   & $1.69$  & $1.28$ & $27.94$  &   $+0.30$\\
    3   &   ~~ + Linear Combination		      & $+14.7$M	& $1.59$  & $1.26$	& $28.09$ &   $+0.45$\\
    4   &   ~~ + Iterative Aggregation 	      &$+31.5$M	&$1.32$  &  $1.22$	& $28.61$ &   $+0.97$\\
    %~~~~~~~~ + $\mathcal{L}_{diversity}$  &$+31.5$M	& $1.28$  &	&   24.62    &   28.76 \\
    5   &   ~~ + Hierarchical Aggregation 	        &$+23.1$M	&$1.46$  &$1.25$	& $28.63$   &   $+0.99$\\
    %~~~~~~~~ + $\mathcal{L}_{diversity}$    &   $+23.1$M	& $1.41$  &	&   \bf 24.76    &   \bf 28.78 \\
    \hdashline
    6   &   ~~ + Multi-Layer Attention (k=2)		        &$+33.6$M	& $1.19$  &	$1.05$ &    $28.58$   & $+0.94$\\
    7   &   ~~ + Multi-Layer Attention (k=3)		        &$+37.8$M	& $1.12$  &	$1.00$ &    $28.62$   & $+0.98$\\
    \hline
    8   &   ~~ + Iterative Aggregation + $\mathcal{L}_{diversity}$  &$+31.5$M	& $1.28$  &	$1.22$ &   $28.76$   &   $+1.12$\\
    9   &   ~~ + Hierarchical Aggregation + $\mathcal{L}_{diversity}$    &   $+23.1$M	& $1.41$    & $1.25$	&   $28.78$   &   $+1.14$\\
    10   &   ~~ + Multi-Layer Attention (k=2)+ $\mathcal{L}_{diversity}$   &  $+33.6$M	& $1.12$	&  $1.05$ & $28.75$    &  $+1.11$  \\
  \end{tabular}
  \caption{Evaluation of translation performance on WMT14 English$\Rightarrow$German (``En$\Rightarrow$De'') translation task. ``\# Para.'' denotes the number of parameters, and ``Train'' and ``Decode'' respectively denote the training speed (steps/second) and decoding speed (sentences/second) on Tesla P40.}
  \label{tab:main}
\end{table*}

\begin{table*}[t]
  \centering
  \begin{tabular}{l|l||rl|rl}%|c}
    \multirow{2}{*}{\bf System}  &   \multirow{2}{*}{\bf Architecture}  &  \multicolumn{2}{c|}{\bf En$\Rightarrow$De}  & \multicolumn{2}{c}{\bf Zh$\Rightarrow$En}\\
    \cline{3-6}
        &   &   \# Para.    &   BLEU    &   \# Para.    &   BLEU\\
    \hline \hline
    \multicolumn{6}{c}{{\em Existing NMT systems}} \\
    \hline
    \cite{wu2016google} &   \textsc{Rnn} with 8 layers           &   N/A &   $26.30$  &  N/A &  N/A  \\ 
    \cite{Gehring:2017:ICML}  &   \textsc{Cnn} with 15 layers  &   N/A &   $26.36$  &  N/A &  N/A  \\
    \hline
    \multirow{2}{*}{\cite{Vaswani:2017:NIPS}} &   \textsc{Transformer-Base}    &  65M &   $27.3$  &    N/A & N/A \\ 
    &  \textsc{Transformer-Big}               &  213M &  $28.4$  &  N/A  &  N/A\\ 
    \hdashline
    \cite{hassan2018achieving}  &   \textsc{Transformer-Big}  &  N/A  & N/A &  N/A  &  $24.2$\\
    %\textsc{RNNSearch} + Ensemble \cite{wang2017sogou} & 22.9 (dev) & - \\ 
    \hline\hline
    \multicolumn{6}{c}{{\em Our NMT systems}}   \\ \hline
    \multirow{4}{*}{\em this work}  &   \textsc{Transformer-Base}  &  88M  &   $27.64$   &    108M  & $24.13$\\
    &   ~~ + Deep Representations   &    111M  & $28.78^{\dag}$   &    131M &   $24.76^{\dag}$\\
    \cline{2-6}
    &   \textsc{Transformer-Big}    & 264M  &   $28.58$      &  304M    & $24.56$\\
    &   ~~ + Deep Representations   &  356M & $29.21^{\dag}$   &  396M &    $25.10^{\dag}$\\
  \end{tabular}
  \caption{Comparing with existing NMT systems on WMT14 English$\Rightarrow$German and WMT17 Chinese$\Rightarrow$English tasks. ``+ Deep Representations'' denotes ``+ Hierarchical Aggregation + $\mathcal{L}_{diversity}$''. ``$\dag$'' indicates statistically significant difference ($p < 0.01$) from the \textsc{Transformer} baseline.}
  %Translation qualities are evaluated on test set.} %``\# Param." denotes the parameter size of each Chinese-English translation model.}
  \label{tab:exsit}
\end{table*}

\subsection{Setup}

\paragraph{Dataset.}
To compare with the results reported by previous work~\cite{Gehring:2017:ICML,Vaswani:2017:NIPS,hassan2018achieving}, we conducted experiments on both Chinese$\Rightarrow$English (Zh$\Rightarrow$En) and English$\Rightarrow$German (En$\Rightarrow$De) translation tasks.
For the Zh$\Rightarrow$En task, we used all of the available parallel data with maximum length limited to 50, consisting of about $20.62$ million sentence pairs. We used newsdev2017 as the development set and newstest2017 as the test set. 
For the En$\Rightarrow$De task, we trained on the widely-used WMT14 dataset consisting of about $4.56$ million sentence pairs.  We used newstest2013 as the development set and newstest2014 as the test set.
%All the data had been tokenized and segmented into subword symbols using byte-pair encoding (BPE) \cite{sennrich2016neural}. We learned a BPE model with 32K merge operations for both datasets. The Chinese data had been tokenized using {\it Jieba}~\footnote{https://github.com/fxshy/jieba}. English sentences were tokenized using the scripts provided in Moses. 
%The Zh$\Rightarrow$En corpus consists of 20M sentence pairs, and the En$\Rightarrow$De corpus consists of 4M sentence pairs. We followed previous work to select the validation and test sets.
Byte-pair encoding (BPE) was employed to alleviate the Out-of-Vocabulary problem \cite{sennrich2016neural} with 32K merge operations for both language pairs.
We used 4-gram NIST BLEU score \cite{papineni2002bleu} as the evaluation metric, and {\em sign-test} \cite{Collins05} to test for statistical significance.

\paragraph{Models.}
We evaluated the proposed approaches on advanced Transformer model~\cite{Vaswani:2017:NIPS}, and implemented on top of an open-source toolkit -- THUMT~\cite{zhang2017thumt}. We followed~\newcite{Vaswani:2017:NIPS} to set the configurations and train the models, and have reproduced their reported results on the En$\Rightarrow$De task.
The parameters of the proposed models were initialized by the pre-trained model.
We tried $k=2$ and $k=3$ for the multi-layer attention model, which allows to attend to the lower two or three layers.

We have tested both \emph{Base} and \emph{Big} models, which differ at hidden size (512 vs. 1024), filter size (2048 vs. 4096) and the number of attention heads (8 vs. 16).\footnote{Here ``filter size'' refers to the hidden size of the feed-forward network in the Transformer model.}
All the models were trained on eight NVIDIA P40 GPUs where each was allocated with a batch size of 4096 tokens.
In consideration of computation cost, we studied model variations with \emph{Base} model on En$\Rightarrow$De task, and evaluated overall performance with \emph{Big} model on both Zh$\Rightarrow$En and En$\Rightarrow$De tasks.

\subsection{Results}

Table \ref{tab:main} shows the results on WMT14 En$\Rightarrow$De translation task. 
As seen, the proposed approaches improve the translation quality in all cases, although there are still considerable differences among different variations.

\paragraph{Model Complexity}
Except for dense connection, all other deep representation strategies introduce new parameters, ranging from 14.7M to 33.6M. Accordingly, the training speed decreases due to more efforts to train the new parameters.
Layer aggregation mechanisms only marginally decrease decoding speed, while multi-layer attention decreases decoding speed by 21\% due to an additional attention process for each layer.

\paragraph{Layer Aggregation} (Rows 2-5):
Although dense connection and linear combination only marginally improve translation performance, iterative and hierarchical aggregation strategies achieve more significant improvements, which are up to +0.99 BLEU points better than the baseline model. This indicates that deep aggregations outperform their shallow counterparts by incorporating more depth and sharing, which is consistent with the results in computer vision tasks~\cite{Yu:2018:CVPR}.

\paragraph{Multi-Layer Attention} (Rows 6-7): 
Benefiting from the power of attention models, multi-layer attention model can also significantly outperform baseline, although it only attends to one or two additional layers. However, increasing the number of lower layers to be attended from $k=2$ to $k=3$ only gains marginal improvement, at the cost of slower training and decoding speeds. In the following experiments, we set set $k=2$ for the multi-layer attention model.

\paragraph{Layer Diversity} (Rows 8-10):
The introduced diversity regularization consistently improves performance in all cases by encouraging different layers to capture diverse information. Our best model outperforms the vanilla Transformer by +1.14 BLEU points. In the following experiments, we used hierarchical aggregation with diversity regularization (Row 8) as the default strategy.

\paragraph{Main Results}
Table~\ref{tab:exsit} lists the results on both WMT17 Zh$\Rightarrow$En and WMT14 En$\Rightarrow$De translation tasks. As seen, exploiting deep representations consistently improves translation performance across model variations and language pairs, demonstrating the effectiveness and universality of the proposed approach.
It is worth mentioning that \textsc{Transformer-Base} with deep representations exploitation outperforms the vanilla \textsc{Transformer-Big} model, with only less than half of the parameters.

\subsection{Analysis}

We conducted extensive analysis from different perspectives to better understand our model. %in terms of its effect on different NMT components, impact of different intermediate function choices, as well as building the ability of handling long sentences.
All results are reported on the En$\Rightarrow$De task with \textsc{Transformer-Base}.

\subsubsection{Length Analysis}

\begin{figure}[h]
\centering
\includegraphics[width=0.35\textwidth]{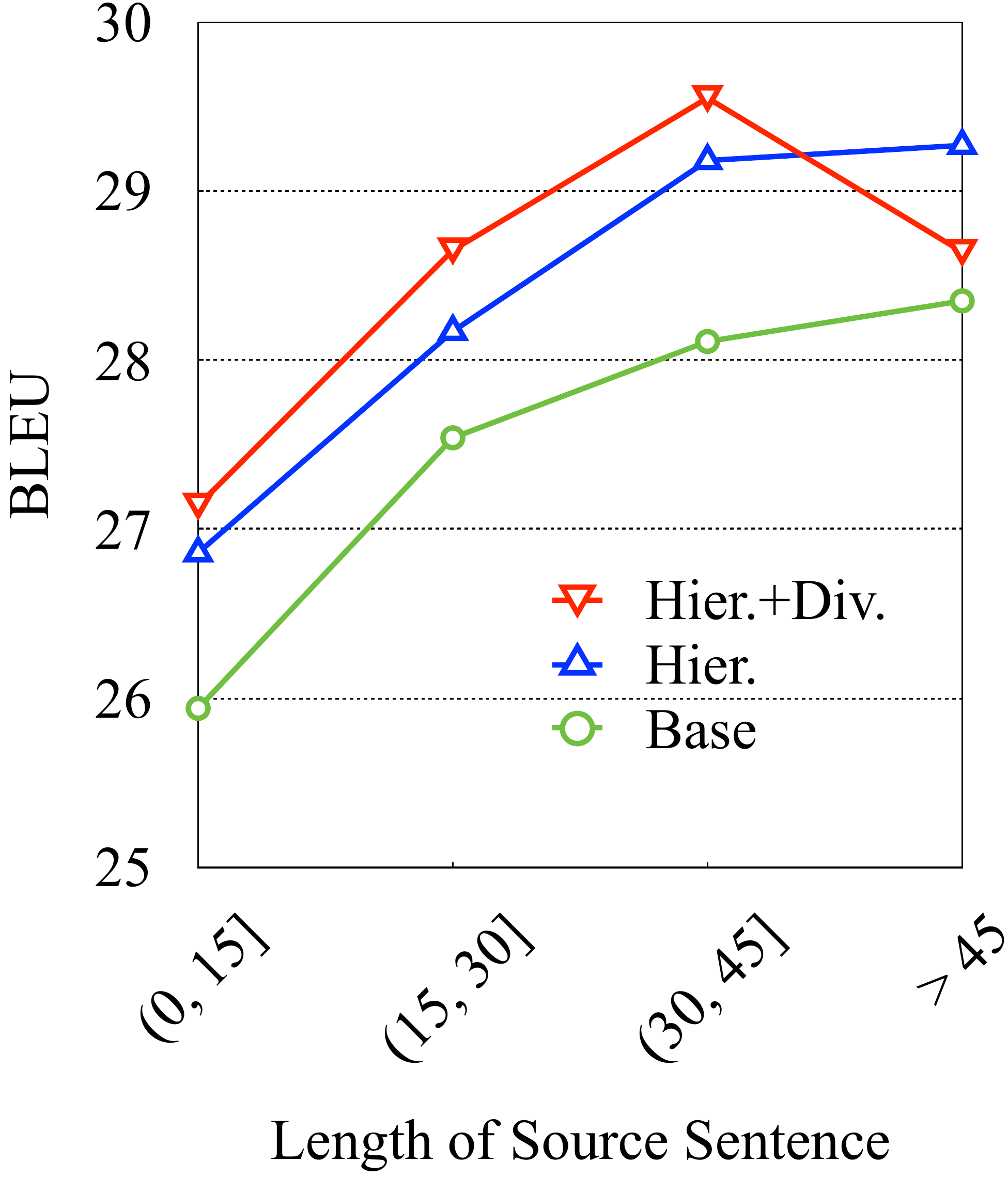}
\caption{BLEU scores on the En$\Rightarrow$De test set with respect to various input sentence lengths. ``Hier.'' denotes hierarchical aggregation and ``Div.'' denotes diversity regularization.}
\label{fig:length}
\end{figure}

Following~\newcite{Bahdanau:2015:ICLR} and ~\newcite{tu2016modeling}, we grouped sentences of similar lengths together and computed the BLEU score for each group, as shown in Figure~\ref{fig:length}.
Generally, the performance of \textsc{Transformer-Base} goes up with the increase of input sentence lengths, which is superior to the performance of RNN-based NMT models on long sentences reported by~\cite{Bentivogli:2016:EMNLP}. We attribute this to the strength of self-attention mechanism to model global dependencies without regard to their distance.

Clearly, the proposed approaches outperform the baseline model in all length segments, while there are still considerable differences between the two variations.
Hierarchical aggregation consistently outperforms the baseline model, and the improvement goes up on long sentences. One possible reason is that long sentences 
%that contain complex syntax and semantic %dependencies, 
indeed require deep aggregation mechanisms.
Introducing diversity regularization further improves performance on most sentences (e.g. $\leq 45$), while the improvement degrades on long sentences (e.g. $>45$). We conjecture that complex long sentences may need to store duplicate information across layers, which conflicts with the diversity objective.

\subsubsection{Effect on Encoder and Decoder}

\begin{table}[h]
  \centering
  \begin{tabular}{c|c|c||c}
    \multirow{2}{*}{\bf Model}   & \multicolumn{2}{c||}{\bf Applied to}      &   \multirow{2}{*}{\bf BLEU}\\  
    \cline{2-3}
                    &       \em Encoder &   \em Decoder   &   \\
    \hline
    \textsc{Base}    &   N/A  &   N/A  &    26.13\\
    \hline
    \hline
    \multirow{3}{*}{\textsc{Ours}}		&   \checkmark  &   \texttimes   & 26.32 \\
    		                            &	\texttimes  &	\checkmark   & 26.41 \\
     		                            &   \checkmark  &   \checkmark   & \bf 26.69 \\
  \end{tabular}
  \caption{Experimental results of applying hierarchical aggregation to different components on  En$\Rightarrow$De validation set.}
  \label{tab:component}
\end{table}

Both encoder and decoder are composed of a stack of $L$ layers, which may benefit from the proposed approach. In this experiment, we investigated how our models affect the two components, as shown in Table~\ref{tab:component}. Exploiting deep representations of encoder or decoder individually consistently outperforms the vanilla baseline model, and exploiting both components further improves the performance. These results provide support for the claim that exploiting deep representations is useful for both understanding input sequence and generating output sequence.

\subsubsection{Impact of Aggregation Choices}

\begin{table}[h]
  \centering
  \scalebox{0.95}{
  \begin{tabular}{c|c|c||c}
    {\bf Model}   &     \bf \textsc{Residual}    &     {\bf \textsc{Aggregate}}   &   {\bf BLEU}\\  
    \hline
    \textsc{Base}    &   N/A    &   N/A     & 26.13\\
    \hline
    \hline
    %\hline
    \multirow{5}{*}{\textsc{Ours}}		&   None        &   \multirow{3}{*}{\textsc{Sigmoid}}     & 25.48\\
                                        &   Top         &        & 26.59\\
                                        \cline{2-2}
                                        &   \multirow{3}{*}{All}    &        & \bf 26.69 \\
                                        \cline{3-3}
    		                            &       &	\textsc{ReLU}        & 26.56 \\
     		                            &       &   \textsc{Attention}    & 26.54 \\
  \end{tabular}
  }
  \caption{Impact of residual connections and aggregation functions for hierarchical layer aggregation. }
  \label{tab:aggregation}
\end{table}

As described in Section~\ref{sec:aggregation}, the function of hierarchical layer aggregation is defined as
\begin{equation}
      \textsc{Agg}(x,y,z) = \textsc{Ln}(\textsc{Ffn}([x;y;z]) + x + y + z), \nonumber
\end{equation}
where $\textsc{Ffn}(\cdot)$ is a feed-forward network with a sigmoid activation in between. 
In addition, all the input layers $\{x, y, z\}$ have residual connections to the output. 
In this experiment, we evaluated the impact of residual connection options, as well as different choices for the aggregation function, as shown in Table~\ref{tab:aggregation}.

Concerning residual connections, if none of the input layers are connected to the output layer (``None''), the performance would decrease.
The translation performance is improved when the output is connected to only the top level of the input layers (``Top''), while connecting to all input layers (``All'') achieves the best performance.
This indicates that cross-layer connections are necessary to avoid the gradient vanishing problem.

Besides the feed-forward network with sigmoid activation, we also tried two other aggregation functions for $\textsc{Ffn}(\cdot)$: (1)  
A feed-forward network with a \textsc{ReLU} activation in between; and (2) multi-head self-attention layer that constitutes the encoder and decoder layers in the \textsc{Transformer} model. As seen, all the three functions consistently improve the translation performance, proving the robustness of the proposed approaches.

\subsection{Visualization of Aggregation}

\begin{figure}[h]
\centering
\subfloat[\small No regularization.]{
    \includegraphics[width=0.2\textwidth]{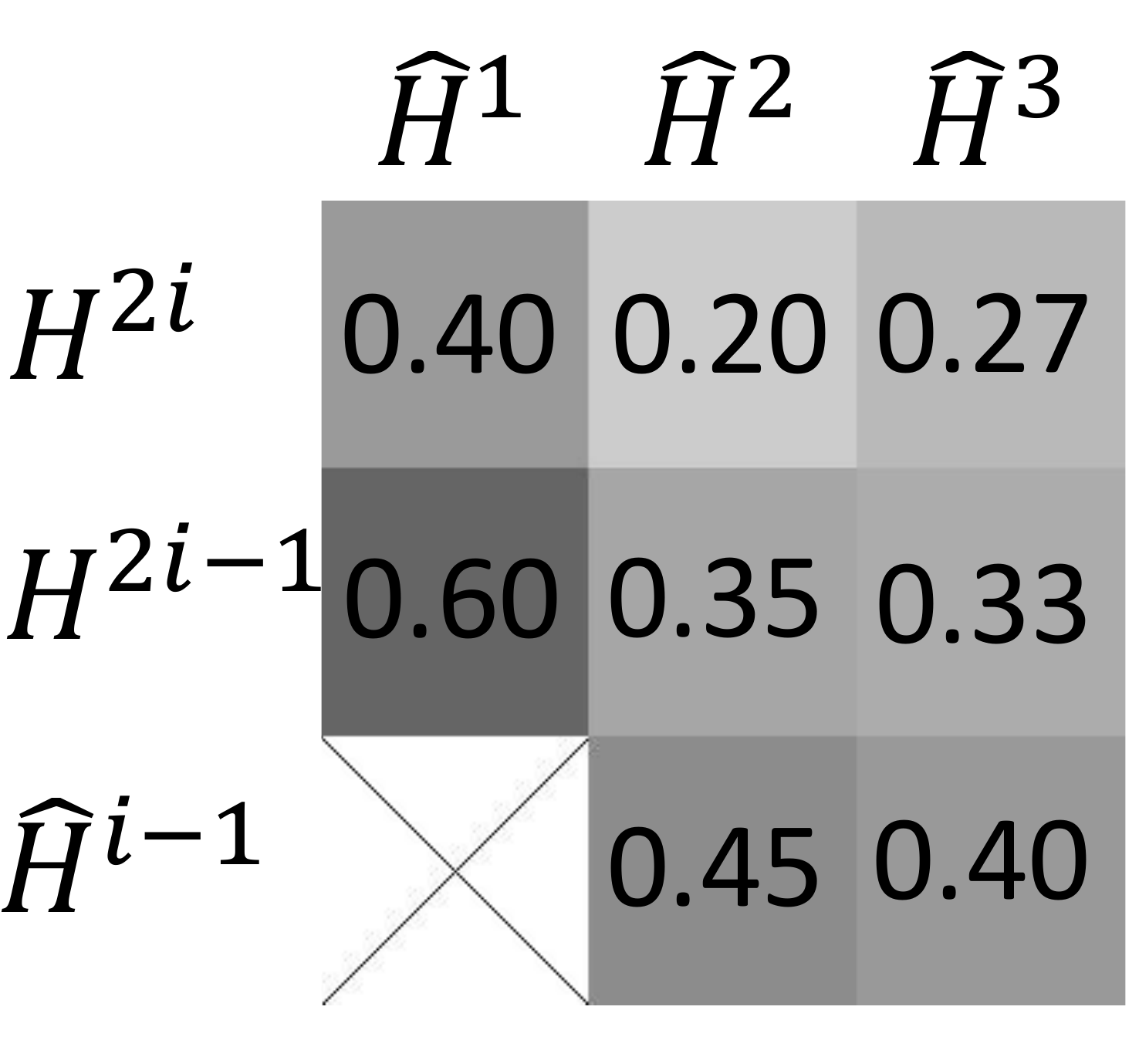}} \hfill
\subfloat[\small With regularization.]{
    \includegraphics[width=0.2\textwidth]{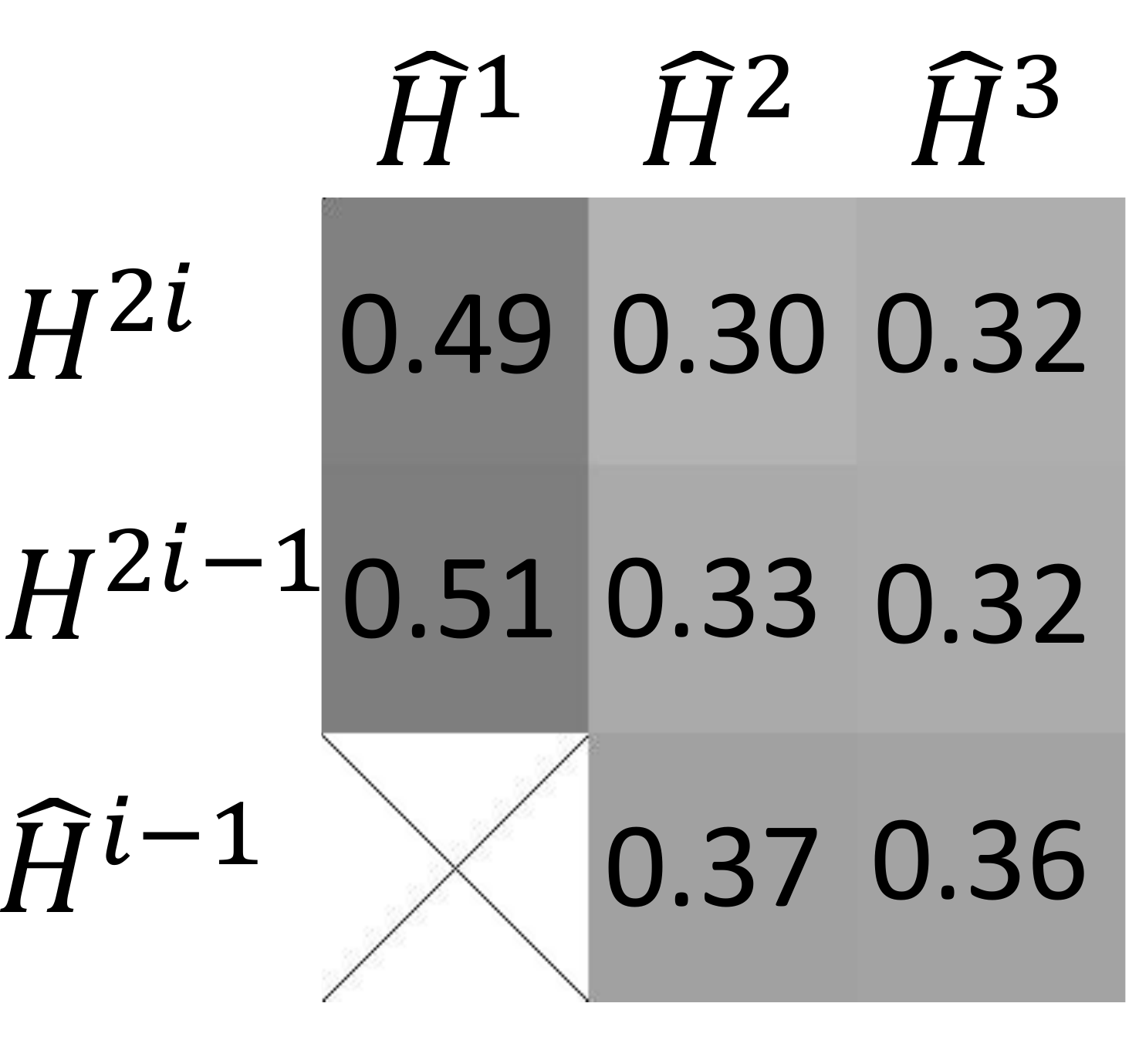}}
\caption{Visualization of the exploitation of input representations for hierarchical aggregation. $x$-axis is the aggregation node and $y$-axis is the input representation. $\widehat{H}^{i}$ denotes the $i$-th aggregation layer, and ${H}^{i}$ denotes the $i$-th encoder layer. The rightmost and topmost position in $x$-axis and $y$-axis respectively represent the highest layer.}
\label{fig:div_effect}
\end{figure}

To investigate the impact of diversity regularization, we visualized the exploitation of the input representations for hierarchical aggregation in encoder side, as shown in Figure~\ref{fig:div_effect}. 
Let ${\bf H}^i = \{H^{2i}, H^{2i-1}, \widehat{H}^{i-1}\}$ be the input representations, we calculated the exploitation of the $j$-th input as 
% the sum of the absolute values of the parameter matrix associated with the input representations, which is normalized by the 
\begin{equation}
    s_j = \frac{\sum_{w \in W_j} |w|}{\sum_{H_{j'} \in {\bf H}}\{\sum_{w' \in W_{j'}} |w'|\}},
\end{equation}
where $W_j$ is the parameter matrix associated with the input $H_j$. 
The score $s_j$ is a rough estimation of the contribution of $H_j$ to the aggregation $\widehat{H}^i$.
% Larger score means more exploitation of the corresponding input.

We have two observations. First, the model tends to utilize the bottom layer more than the top one, indicating the necessity of fusing information across layers. Second, using the diversity regularization in Figure~\ref{fig:div_effect}(b) can encourage each layer to contribute more equally to the aggregation. We hypothesize this is because of the diversity regularization term encouraging the different layers to contain diverse and equally important information.

\section{Related Work}

Representation learning is at the core of deep learning. Our work is inspired by technological advances in representation learning, specifically in the field of {\em deep representation learning} and {\em representation interpretation}.

\paragraph{Deep Representation Learning}
Deep neural networks have advanced the state of the art in various communities, such as computer vision and natural language processing. One key challenge of training deep networks lies in how to transform information across layers, especially when the network consists of hundreds of layers. 

In response to this problem, ResNet~\cite{he2016deep} uses skip connections to combine layers by simple, one-step operations. Densely connected network~\cite{Huang:2017:CVPR} is designed to better propagate features and losses through skip connections that concatenate all the layers in stages. \newcite{Yu:2018:CVPR} design structures iteratively and hierarchically merge the feature hierarchy to better fuse information in a deep fusion.

Concerning machine translation,~\newcite{Meng:2016:ICLRWorkshop} and~\newcite{Zhou:2016:TACL} have shown that deep networks with advanced connecting strategies outperform their shallow counterparts. Due to its simplicity and effectiveness, skip connection becomes a standard component of state-of-the-art %deep 
NMT models~\cite{wu2016google,Gehring:2017:ICML,Vaswani:2017:NIPS}. In this work, we prove that deep representation exploitation can further improve performance over simply using skip connections.

\paragraph{Representation Interpretation}
Several researchers have tried to visualize the representation of each layer to help better understand what information each layer captures~\cite{zeiler2014visualizing,li2016convergent,ding2017visualizing}.
Concerning natural language processing tasks, 
~\newcite{shi-padhi-knight:2016:EMNLP2016} find that both local and global source syntax are learned by the NMT encoder and different types of syntax are captured at different layers.
~\newcite{Anastasopoulos:2018:NAACL} show that higher level layers are more representative than lower level layers.
~\newcite{Peters:2018:NAACL} demonstrate that higher-level layers capture context-dependent aspects of word meaning while lower-level layers model aspects of syntax.
Inspired by these observations, we propose to expose all of these representations to better fuse information across layers. In addition, we introduce a regularization to encourage different layers to capture diverse information.

\section{Conclusion}
In this work, we propose to better exploit deep representations that are learned by multiple layers for neural machine translation.
Specifically, the hierarchical aggregation with diversity regularization achieves the best performance by incorporating more depth and sharing across layers and by encouraging layers to capture different information. 
Experimental results on WMT14 English$\Rightarrow$German and WMT17 Chinese$\Rightarrow$English show that
the proposed approach consistently outperforms the state-of-the-art \textsc{Transformer} baseline by +0.54 and +0.63 BLEU points, respectively.
% translation tasks demonstrate the efficiency and universality of the proposed approach.
By visualizing the aggregation process, we find that our model indeed utilizes lower layers to effectively fuse the information across layers.

Future directions include validating our approach on other architectures such as RNN~\cite{Bahdanau:2015:ICLR} or CNN~\cite{Gehring:2017:ICML} based NMT models, as well as combining with other advanced techniques~\cite{shaw2018self,Shen:2018:AAAI,Yang:2018:EMNLP,Li:2018:EMNLP} to further improve the performance of \textsc{Transformer}.

\section*{Acknowledgments}
We thank the anonymous reviewers for their insightful comments.

%As our approach is not limited to specific architectures, it is interesting to validate our model on other architectures, such as RNN~\cite{wu2016google} and CNN~\cite{Gehring:2017:ICML} based NMT models. Another promising direction is to combine with other advanced techniques~\cite{shaw2018self,Shen:2018:AAAI} to further improve the performance of \textsc{Transformer}.

%Based on the idea that for deep NMT models, different layers would capture diverse information, we propose several techniques to better fuse the information across layers. From simple linear combination of layers to complex, deep aggregation strategies, we conduct extensive experiments to verify the effectiveness of aggregating layers. In addition, we propose a criterion to measure the diversity between layers and testify that adding the diversity regularization can further enhance the performance of models.
%Our best model could enhance the performance of the base and big Transformer model by $+1.14$ and $+0.63$ BLEU points on English$\Rightarrow$German dataset and $+0.63$ and $+0.54$ BLEU points on Chinese$\Rightarrow$English dataset respectively.  

% \balance
\balance
\bibliographystyle{acl_natbib}
\bibliography{ref.bib}
\end{document}